\documentclass[10pt,twocolumn,letterpaper]{article}

\usepackage{tabularx}
\usepackage{booktabs}
\usepackage{amsmath}
\usepackage{amssymb}
\usepackage{amsfonts}
\usepackage{amsmath,amsfonts,amssymb,amsthm,dsfont}
\usepackage{graphicx}
\usepackage{cite}
\usepackage{todonotes}
\usepackage{paralist}
\usepackage{array}
\usepackage{multirow}
\usepackage[sort, numbers]{natbib}

\usepackage[pagebackref,breaklinks,colorlinks]{hyperref}
\usepackage{url}

\usepackage[capitalize]{cleveref}
\crefname{section}{Sec.}{Secs.}
\Crefname{section}{Section}{Sections}
\Crefname{table}{Table}{Tables}
\crefname{table}{Tab.}{Tabs.}

\def\R{\mathbb{R}}
\def\nor{\mathcal{N}}
\def\F{\mathcal{F}}
\def\W{\mathcal{W}}
\def\Z{\mathcal{Z}}

\def\our{PluGeN4Faces}
\def\plug{PluGeN}

\date{}

\usepackage[format=plain,labelformat=simple,labelsep=period,font=small,compatibility=false]{caption}
\usepackage[font=footnotesize,skip=3pt,subrefformat=parens]{subcaption}
\makeatletter
\DeclareRobustCommand\onedot{\futurelet\@let@token\@onedot}
\def\@onedot{\ifx\@let@token.\else.\null\fi\xspace}

\makeatother

\begin{document}

%%%%%%%%% TITLE - PLEASE UPDATE
\title{\textbf{Face Identity-Aware Disentanglement in StyleGAN}}

\author{
Adrian~Suwała\\
% Faculty of Mathematics and Computer Science,\\ Jagiellonian University\\
Jagiellonian University\\
\and
Bartosz~Wójcik\\
% Faculty of Mathematics and Computer Science,\\ Jagiellonian University\\
% Doctoral School of Exact and Natural Sciences,\\Jagiellonian University\\
Jagiellonian University\\
IDEAS NCBR\\
\and
Magdalena~Proszewska\\
University of Edinburgh\\
\and
Jacek~Tabor\\
% Faculty of Mathematics and Computer Science,\\ Jagiellonian University\\
Jagiellonian University\\
\and
Przemysław~Spurek\\
% Faculty of Mathematics and Computer Science,\\ Jagiellonian University\\
Jagiellonian University\\
\and
Marek Śmieja\\
% Faculty of Mathematics and Computer Science,\\ Jagiellonian University\\
Jagiellonian University\\
}
\maketitle

%%%%%%%%% ABSTRACT
\begin{abstract}
Conditional GANs are frequently used for manipulating the attributes of face images, such as expression, hairstyle, pose, or age. Even though the state-of-the-art models successfully modify the requested attributes, they simultaneously modify other important characteristics of the image, such as a person's identity. In this paper, we focus on solving this problem by introducing \our{}, a plugin to StyleGAN, which explicitly disentangles face attributes from a person's identity. Our key idea is to perform training on images retrieved from movie frames, where a given person appears in various poses and with different attributes. By applying a type of contrastive loss, we encourage the model to group images of the same person in similar regions of latent space. Our experiments demonstrate that the modifications of face attributes performed by \our{} are significantly less invasive on the remaining characteristics of the image than in the existing state-of-the-art models.
\end{abstract}

%%%%%%%%% BODY TEXT
\section{Introduction}
Modern generative models, such as StyleGAN \citep{karras2019style,Karras2019stylegan2,karras2021alias}, produce high-quality images, which are frequently indistinguishable from real ones. One of the current challenges is to introduce the functionality for manipulating the attributes of existing images. In the case of face images, we would like to modify the expression, the type of facial hair, or even the gender of the person in the photo. 

Although the state-of-the-art conditional generative models, such as \plug{} \citep{wolczyk2022plugen} or StyleFlow \citep{abdal2020styleflow}, are capable of modifying selected face attributes, there is no guarantee that only requested attributes are changed. Experiments show that modifications of intended attributes often affect other attributes as well as the identity of a person. It means that the latent space used for modifications is so entangled that manipulating only selected attributes independently from other characteristics of the image is impossible.

  \newcolumntype{C}{>{\centering\arraybackslash}X}
  \begin{figure}[ht!]
   \centering
   \begin{tabularx}{\linewidth}{CCCCCC} input & gender & \hspace*{-2mm} glasses & hair & beard & smile\end{tabularx} \\
    \includegraphics[width=\linewidth]{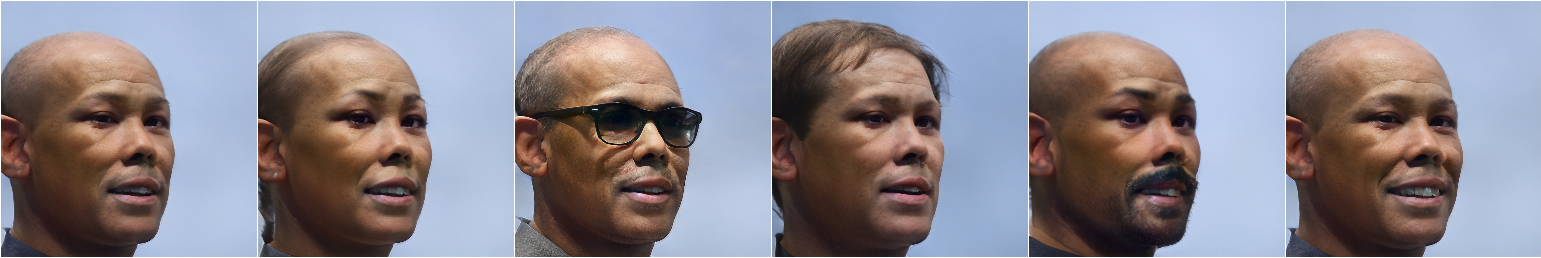}
    \\
    \includegraphics[width=\linewidth]{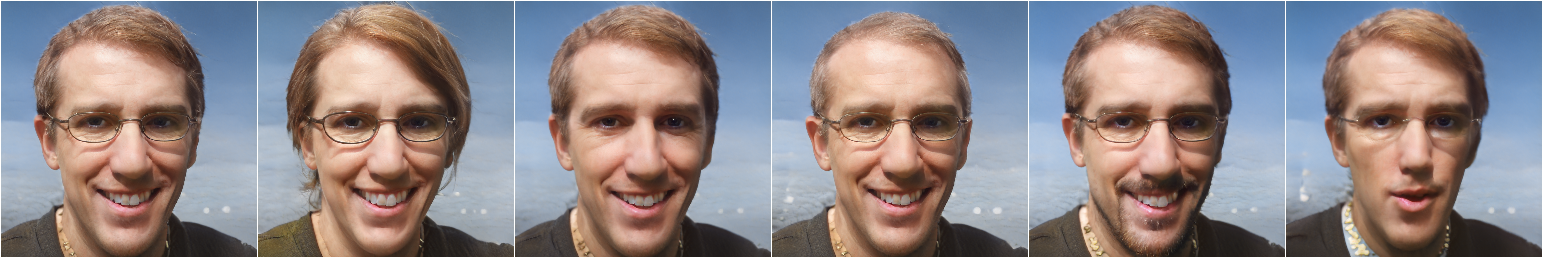} 
    \caption{Sample effects of attributes manipulation performed by \our{}. }
        \label{fig:intro}
 \end{figure}

There may be various reasons why existing models cannot create disentangled latent representation. In this paper, we argue that the conditional generative models are usually trained on generated (fake) images and they have never seen images representing the same person with different combinations of attributes. To introduce the information about the person's identity, we need to perform training on real images instead of generated ones only.

Working with real images is straightforward in autoencoder-based generative models, but there appear notable problems in the case of GANs since there is no built-in method for encoding images into the GAN latent space. The problem is especially challenging for StyleGAN architecture because of the structure of its style space. While generated images are identified by a single style code $\mathbf{w} \in \W 
\subset \R^{512}$, not every image can be accurately mapped into $\W$ \citep{abdal2019image2stylegan}. To overcome this issue, most techniques (employing the encoder or gradient-based optimization) perform the search in the extended style space $\W^k_*$, where a style code consists of $k$ different 512-dimensional style vectors $\mathbf{w}_1,\ldots,\mathbf{w}_k \in \R^{512}$ (typically $k=18$) -- one for each layer of the StyleGAN architecture that can receive input via AdaIn \citep{abdal2019image2stylegan, zhu2020improved, tov2021designing}. Operating on the whole set of style codes significantly increases the dimensionality of latent codes and theoretically makes the complexity of the problem more challenging.

\begin{figure}
    \centering
    \includegraphics[width=0.5\textwidth]{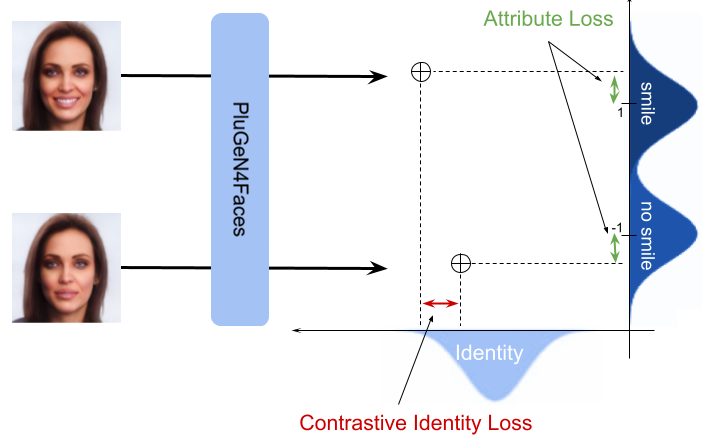}
    \caption{Explicit disentanglement of attribute and identity features performed by \our{}. While each labeled attribute is modeled as an individual latent dimension, the contrastive loss allows us to group latent codes representing images of the same person in similar regions of the space.}
    \label{fig:teaser}
\end{figure}

In this paper, we introduce \our{} ({\bf Plu}gin {\bf Ge}nerative {\bf N}etworks for {\bf Faces}), a plugin model for disentangling the latent space of StyleGAN in the case of face images. \our{} provides full control on manipulating face attributes so that the modification of the requested attributes has a minimal effect on the identity of a person and the remaining face attributes (including background), see Figure \ref{fig:intro} for sample results. \our{} works as a plugin to pre-trained StyleGAN, which means that it does not change the weights of StyleGAN but only transforms its style space into a disentangled one. In consequence, a training process is extremely simple and absorbs limited computational resources.

In contrast to competitive models, \our{} is trained on face images retrieved from movie frames, which can present a given person in various poses and with different attributes. The information about a person's identity is used in \our{} by employing a contrastive loss. Namely, we encourage the model to group images of the same person in similar regions of latent space, see Figure \ref{fig:teaser}. To use real images in training, we implement \our{} as a conditional invertible normalizing flow, where the condition represents the identifier of the style code. In other words, \our{} transforms every style code $\mathbf{w}_i$, for $i=1,\ldots,k$, by the flow conditioned on the index $i$. In this way, we are able to implement a compact disentanglement module operating on real images.%, not only the generated ones as done before. 

We evaluate \our{} on face images retrieved from the FFHQ database as well as movie frames. We show that \our{} allows for effective manipulation of face attributes. Moreover, the applied modifications preserve the person's identity to a significantly greater extent than in competitive models. The presented sample results are supported by the quantitative analysis, which confirms the advantage of \our{} over related models.

The contribution of the paper is summarized as follows:
\begin{compactitem}
    \item We introduce a plugin to StyleGAN for manipulating the attributes of real images. In contrast to existing models, it is trained on real images encoded into StyleGAN style space using the encoder network. 
    \item We improve the representation disentanglement in conditional generative models by applying a type of contrastive loss, which explicitly encodes the person's identity. In consequence, the manipulation of the requested attributes is less invasive on the remaining image characteristics (including person's identity).
    \item The proposed solution is evaluated in a strict quantitative way, which allows for a fair comparison with related models. The proposed metrics together with our sample results clearly demonstrate the advantage of \our{} over competitive methods. 
\end{compactitem}

\section{Related work}

Conditional VAE (cVAE) is one of the first methods of including additional label information in a generative model \citep{kingma2014semi}, which has been successfully applied in a variety of disciplines including image generation~\citep{sohn2015learning,yan2016attribute2image,klys2018learning}.
However, the independence of latent codes and labels is not assured, which has a negative impact on the generation quality. 
Conditional GAN (cGAN) is an alternative that is able to produce examples of significantly better quality~\citep{mirza2014conditional,perarnau2016invertible,he2019attgan, park2019semantic, choi2020stargan}
, but the training of the model is more difficult~\citep{kodali2017convergence}.
%cGAN is able to generate new high-quality images and includes conditioning factors in various forms (images, sketches, labels) \citep{park2019semantic, choi2020stargan}. However, it lacks the ability to manipulate existing examples.
Fader Networks~\citep{lample2017fader} overcome this limitation by combining components of cVAE and cGAN, as they use both encoder-decoder architecture and the discriminator, which predicts the image attributes from a corresponding latent vector obtained from the encoder. As with previous methods, Fader Networks does not preserve the disentanglement of attributes, moreover, the training is even more difficult than that of standard GANs.
%CAGlow~\citep{liu2019conditional} is a vastly different approach, in which Glow~\citep{kingma2018glow} is used for conditional image generation by modeling a joint probabilistic density of an image and its attributes. However, it does not reduce data dimension, therefore is not as applicable to more complex data.

While the described approaches focus on creating conditional generative models from scratch, recent work frequently focuses on manipulating the latent codes of pre-trained networks. In this scenario, data complexity is not that big of a limitation, hence flow models can be easily applied. 
StyleFlow~\citep{abdal2020styleflow} and \plug{}~\citep{wolczyk2022plugen} operate on the latent space of GAN using a normalizing flow module: conditional CNF~\citep{grathwohl2018ffjord} and NICE~\citep{dinh2015nice}, respectively. While StyleFlow is adapted to work only on StyleGAN \citep{Karras2019stylegan2}, \plug{} demonstrates great results also with other models and in different domains. For StyleGAN, they are both trained using latent codes sampled from latent space $W$ and attributes of images that correspond to them. %Our work can be seen as an extension of \plug{}, which provide better disentanglement between attributes and identity in the case of face images.
Competitive approaches include \citep{gao2021high, tewari2020pie, harkonen2020ganspace, nitzan2020disentangling}. InterFaceGAN \citep{shen2020interfacegan} aims to manipulate various properties of the facial semantics via linear models applied to the latent space of GANs. Hijack-GAN~\citep{wang2020hijack} goes beyond linear models and designs a proxy model to traverse the latent space of GANs.

Along with the latent codes manipulation techniques, methods of embedding examples into the GAN latent space can be used to allow manipulation of existing examples. %Specifically, embedding images into the StyleGAN latent space. 
There are two main embedding approaches: (i) an encoder network that maps an image into the latent space \citep{tov2021designing}, (ii) an optimization algorithm that iteratively improves a latent code so that it produces a desired image \citep{abdal2019image2stylegan,abdal2020image2stylegan,zhu2021improved}. Moreover, combinations of these two approaches exist, in which the encoder outputs an approximate embedding that is then improved by the optimization algorithm \citep{zhu2020indomain}.
These methods allow us to train our model using real images, which are encoded into the extended StyleGAN latent space $\W^k_*$ that enables manipulation of existing images. As shown in \citep{abdal2019image2stylegan}, the use of $\W^k_*$ latent space instead of $\W$ reduces the alteration of the original image.

% Disentangled representation learning is a technique of disentangling each feature into variables encoded as separate dimensions. It enables making changes of one factor (corresponding dimension) while being relatively invariant to changes of other factors~\citep{kim2018disentangling,higgins2016beta,chen2019isolating,sorrenson2020disentanglement,chen2016infogan}. 
%\citep{bengio2013representation,kim2018disentangling,higgins2016beta,brakel2017learning,kumar2017variational,chen2019isolating,spurek2020non,dinh2014nice,sorrenson2020disentanglement,chen2016infogan,wolczyk2022plugen}. 
% In this paper, we further investigate \plug{} method \citep{wolczyk2022plugen} and aim to improve disentanglement by training on face images retrieved from movie frames and employing a contrastive loss.

\section{Identity-aware disentanglement}

\paragraph{Overview} 

\our{} is a conditional invertible normalizing flow module (cINF), which is attached to the style space of StyleGAN. It transforms the style codes of pre-trained StyleGAN into a disentangled space so that:
\begin{compactitem}
    \item the labeled attributes are modeled by the individual latent coordinates, 
    \item images of the same person are grouped in similar regions of the latent space.
\end{compactitem}
While realizing the first of the above conditions allows us to edit the values of requested attributes, the second one prevents severe changes in the image during attribute manipulation.

In this section, we first review the StyleGAN architecture and recall the way of encoding real images into its style space. Next, we present a probabilistic structure of \our{}, and cINF mapping function. We discuss the training procedure and the inference phase. %We emphasize the role of contrastive loss, which is applied to images retrieved from movie frames to force explicit encoding of the face identity.

% In this section, we introduce \our{}. First, we recall the basic idea of representing images in the StyleGAN latent space and explain how \plug{} approaches to disentangle the latent space of generative models. Based on these facts, we construct \our{}, which, in addition to modeling labeled attributes, allows us to explicitly control the face identity.

\paragraph{StyleGAN architecture}

StyleGAN architecture \citep{karras2019style} consists of two main parts: (a) a mapping network that transforms latent codes $\mathbf{z} \in \Z$ sampled from Gaussian noise $\nor(\mu,I)$ to the style vectors $\mathbf{w} \in \W$, (b) a synthesis network that creates an image from the style code replicated several times. The replicated style codes represent the inputs to subsequent layers of the synthesis network.

Instead of manipulating latent codes $\mathbf{z} \in \Z$, we usually operate on the style space $\W$ to perform attribute modification, which was shown to be significantly more disentangled \citep{abdal2020styleflow}. However, it is well-known that not all real images can be encoded into the StyleGAN's style space $\W$ \citep{zhu2020improved}. %This problem appears even for images from the training data distribution. 
A typical  approach for coping with this issue is to extend the search space and look for $k$ different style codes $(\mathbf{w}_1,\ldots,\mathbf{w}_k) \in \W^k_*$, which together could synthesize the original input 
\citep{abdal2019image2stylegan, tov2021designing}. Each $\mathbf{w}_i$ represents the input to the $i$-th layer of the synthesis network. Even though a sequence of style codes from the extended style space does not reflect any latent code $\mathbf{z}$, it allows for the convenient reconstruction and manipulation of real images. One can design an encoder \citep{tov2021designing} or implement a gradient-based procedure for embedding real images into the extended style space. In this paper, we employ an encoder network.

% In this paper, we employ the encoder network for embedding real images into the extended style space of StyleGAN.

\paragraph{Probabilistic structure of \our{}}

We assume that every image $\mathbf{x}$ is described by the composition of the attribute and non-attribute vectors $(\mathbf{c},\mathbf{s})$, where $\mathbf{c} \in \mathbf{C}=(C_1,\ldots,C_M)$ and  $\mathbf{s} \in \mathbf{S}=(S_1,\ldots,S_{N-M})$. While each attribute variable $c_i \in C_i$ contains information about the selected attribute, the non-attribute vector $\mathbf{s}$ is used to describe the remaining characteristic of data including background and personal identity in the case of face images. To control the value of every attribute independently of each other, a factorized form of the probability distribution of the random vector $(\mathbf{C},\mathbf{S})$ is assumed. Given a vector of true labels  $\mathbf{y}=(y_1,\ldots,y_M)$, the conditional distribution of $(\mathbf{c},\mathbf{s})$ is defined by
$$
p_{\mathbf{C},\mathbf{S}|\mathbf{Y}=\mathbf{y}}(\mathbf{c},\mathbf{s}) = \prod_{i=1}^M p_{C_i|Y_i=y_i}(c_i) \cdot p_{\mathbf{S}}(\mathbf{s}) \text{ , for } (\mathbf{c},\mathbf{s}) \in \R^N.
$$
In the above formula, the $i$-th label $y_i$ affects only the $i$-th attribute variable $C_i$. As a parametric form of $p_{C_i|Y_i=y_i}$, we use a 1-dimensional Gaussian density $\nor(y_i,\sigma)$. By changing the condition $Y_i=y_i$, we modify the mean of the Gaussian. The distribution of the non-attribute vector is modeled as a multivariate standard Gaussian density $\nor(\mathbf{0},\mathbf{I}_{N-M})$. The non-attribute vector $\mathbf{s}$ is responsible for covering information about a person's identity, image background, etc., so images presenting the same person should have similar values of $\mathbf{s}$. %This construction is analogical to \citep{wolczyk2022plugen}.

\begin{figure}[!htb]
    \centering
    \includegraphics[width=0.5\textwidth]{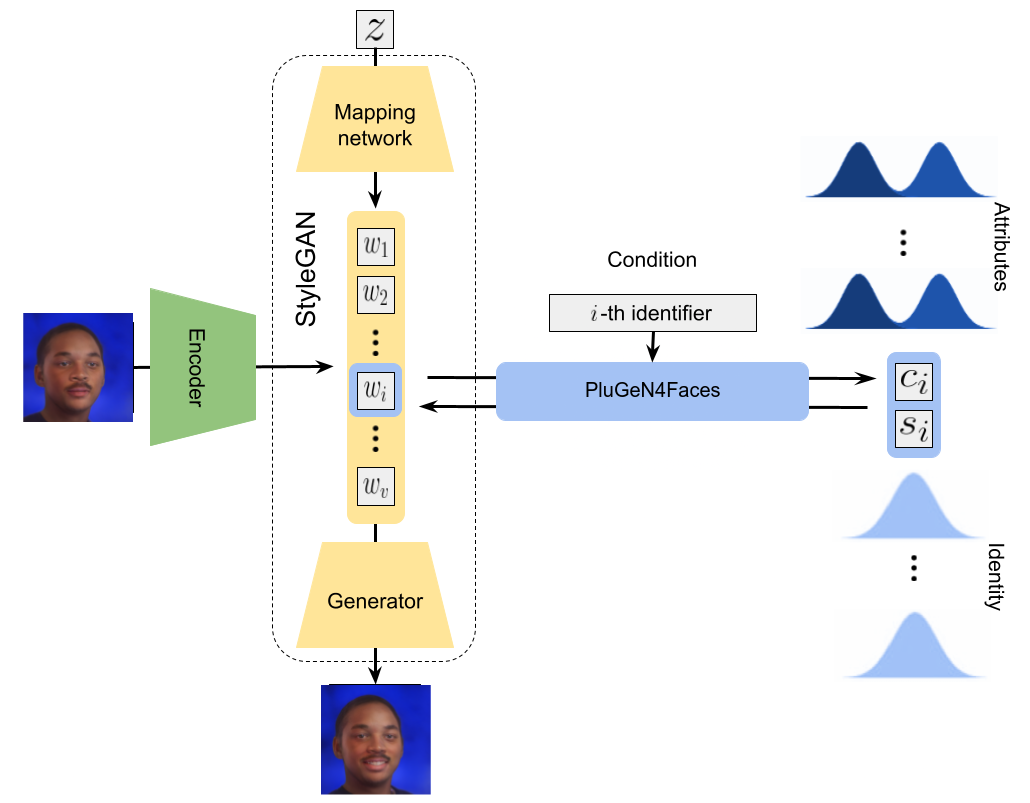}
    \caption{Architecture of \our{}. Given the representation of the input image using a sequence of style codes, \our{} uses INF to model labeled attributes as individual latent dimensions. The remaining characteristic of the image (including the person's identity) are modeled in separate dimensions using contrastive loss.}
    \label{fig:arch}
\end{figure}

\paragraph{Invertible mapping}

To realize the above parameterization, a two-way mapping between the style space of the pre-trained StyleGAN and the disentangled space $(\mathbf{C},\mathbf{S})$ has to be established. Since we work with real images (not only generated ones), we employ the StyleGAN encoder \citep{tov2021designing}, which produces a sequence of style codes $\{\mathbf{w}_1,\ldots,\mathbf{w}_k\} \in \W_*^k$ representing a given image $\mathbf{x}$ in the subsequent layers of the StyleGAN synthesis network. Thus we need to map a sequence of style codes $(\mathbf{w}_i)_{i=1}^k$ (representing a single image $\mathbf{x}$) into a sequence of the attribute and non-attribute vectors $(\mathbf{c}_i, \mathbf{s}_i)_{i=1}^k$. To find such an invertible transformation, we use a conditional INF (cINF), which is parametrized by the identifier of the style code. More precisely, the cINF, $\F: \R^N \to \R^N$, takes the style code $\mathbf{w}_i$ and the index of $i$-th layer as a condition and returns a disentangled representation of $\mathbf{w}_i$ as:
$$
(\mathbf{c}_i,\mathbf{s}_i) = \F(\mathbf{w}_i | \mathrm{layer} = i).
$$
Here, both $\mathbf{c}_i = (c^i_{1},\ldots,c^i_{M})$ and $\mathbf{s}_i=(s^i_{1},\ldots,s^i_{N-M})$ are vectors corresponding to a given $\mathbf{w}_i$. 

% Due to the invertibility, we can transform latent codes $\mathbf{w}$ to attribute-style vectors $(\mathbf{c},\mathbf{s})$, perform the requested modifications on $\mathbf{c}$, and map the resulting vector back $\W$.

\paragraph{Training}

The conditional INF is trained by minimizing the negative log-likelihood taken over all style codes. Given a sequence of style codes  $(\mathbf{w}_i)_{i=1}^k$ representing an image $\mathbf{x}$ with labels $\mathbf{y}$, we aim at minimizing:
\begin{multline} \label{eq:condlogloss}
-\sum_{i=1}^k \log p_{\mathbf{W}_i|\mathbf{Y}= \mathbf{y}}(\mathbf{w}_i) =\\
- \sum_{i=1}^k \log \left(p_{\mathbf{C},\mathbf{S} | \mathbf{Y} = \mathbf{y}}(\mathbf{c}_i,\mathbf{s}_i) \cdot \left|\det \frac{\partial \F^{-1}(\mathbf{w}_i | \mathrm{layer}=i)}{\partial \mathbf{w_i}}\right| \right)=\\
 -\sum_{i=1}^k \left(\sum_{j=1}^M\log p_{C_i^j|Y_i=y_i}(c_i^j) + \log p_{\mathbf{S}_i}(\mathbf{s}_i) + \right. \\
 \left. \log \left|\det \frac{\partial \F^{-1}(\mathbf{w}_i|\mathrm{layer}=i)}{\partial \mathbf{w}_i}\right|\right),
\end{multline}
where $(\mathbf{c}_i,\mathbf{s}_i) = \F^{-1}(\mathbf{w}_i|\mathrm{layer}=i)$ are the attribute and non-attribute vectors describing the $i$-th style code $\mathbf{w}_i$ (in the $i$-th StyleGAN layer).

In addition to the negative log-likelihood minimization, which focuses on modeling labeled attributes, we introduce a contrastive loss responsible for the explicit encoding of the face identity. Thanks to the contrastive loss, manipulating the labeled attributes will have a minimal effect on changing other attributes (including identity) of the face image.

To construct our contrastive loss, we take $n$ images $\mathbf{x}_1,\ldots,\mathbf{x}_n$ of a given person and encode them into the style space of StyleGAN using the encoder network. Such images can be retrieved from subsequent frames of movies. For each image, the encoder produces a sequence of style codes, which represent the input to subsequent layers of StyleGAN generator. For transparency, we restrict our attention to the $l$-th layer in the following description. For $n$ images, we have $n$ style codes $\mathbf{w}_1,\ldots,\mathbf{w}_n$, in which $\mathbf{w}_i$ is the representation of $\mathbf{x}_i$ in the $l$-th layer (we drop the index of the layer for simplicity). Making use of conditional INF, we find a disentangled representation of $\mathbf{w}_i$ as
$$
(\mathbf{c}_i,\mathbf{s}_i) = \F(\mathbf{w}_i|\mathrm{layer} = l).
$$
To force the structure on non-attribute variables, where images of the same person are represented by similar non-attribute vectors, we apply the following contrastive loss:
\begin{equation} \label{eq:contrastive}
\sum_{i \neq j} \|\mathbf{s}_i - \mathbf{s}_j\|^2 = 2n \sum_{i=1}^n \|\mathbf{s}_i - \mathbf{m}\|^2,
\end{equation}
where the mean $\mathbf{m} = \frac{1}{n} \sum_{i=1}^n \mathbf{s}_i$ is used to reduce the number of comparisons \citep{smieja2015spherical}. Minimization of \eqref{eq:contrastive} leads to mapping the set of $n$ input images to similar values of non-attributes vectors. We apply this loss to images representing the same person.

To sum up, the complete loss of \our{} is given by taking together the introduced contrastive loss \eqref{eq:contrastive} and negative log-likelihood \eqref{eq:condlogloss}. For the first loss component, we need a set of images representing the same person, while for the second one, we use images with labeled attributes. %In practice, we can train the model in two ways. The simpler one would be to use a single set of movie frames, where all attributes of each image are labeled. Alternatively, we could use two separate sets -- images from the movies for contrastive loss and standard images with labeled attributes for log-likelihood loss. In the second option, movie images can also have labeled attributes. In this paper, we prefer the latter approach. 

  \newcolumntype{C}{>{\centering\arraybackslash}X}
  % \newcolumntype{s}{\hsize=0.1cm}
  \begin{figure*}[t!]
   \centering
   \begin{tabularx}{0.48\linewidth}{C} {\bf \large Single attribute manipulation} \end{tabularx} \quad \begin{tabularx}{0.48\linewidth}{C} {\bf \large Sequential edits of multiple attributes} \end{tabularx}\\
   \hspace*{1mm}  \small \begin{tabularx}{0.48\linewidth}{CCCCCC} \small  input &  gender & glasses & hair & beard & smile\end{tabularx} \quad \begin{tabularx}{0.48\linewidth}{CCCCCC} \small  input &  gender & glasses & hair & beard & smile\end{tabularx} \\
    \includegraphics[width=0.48\linewidth]{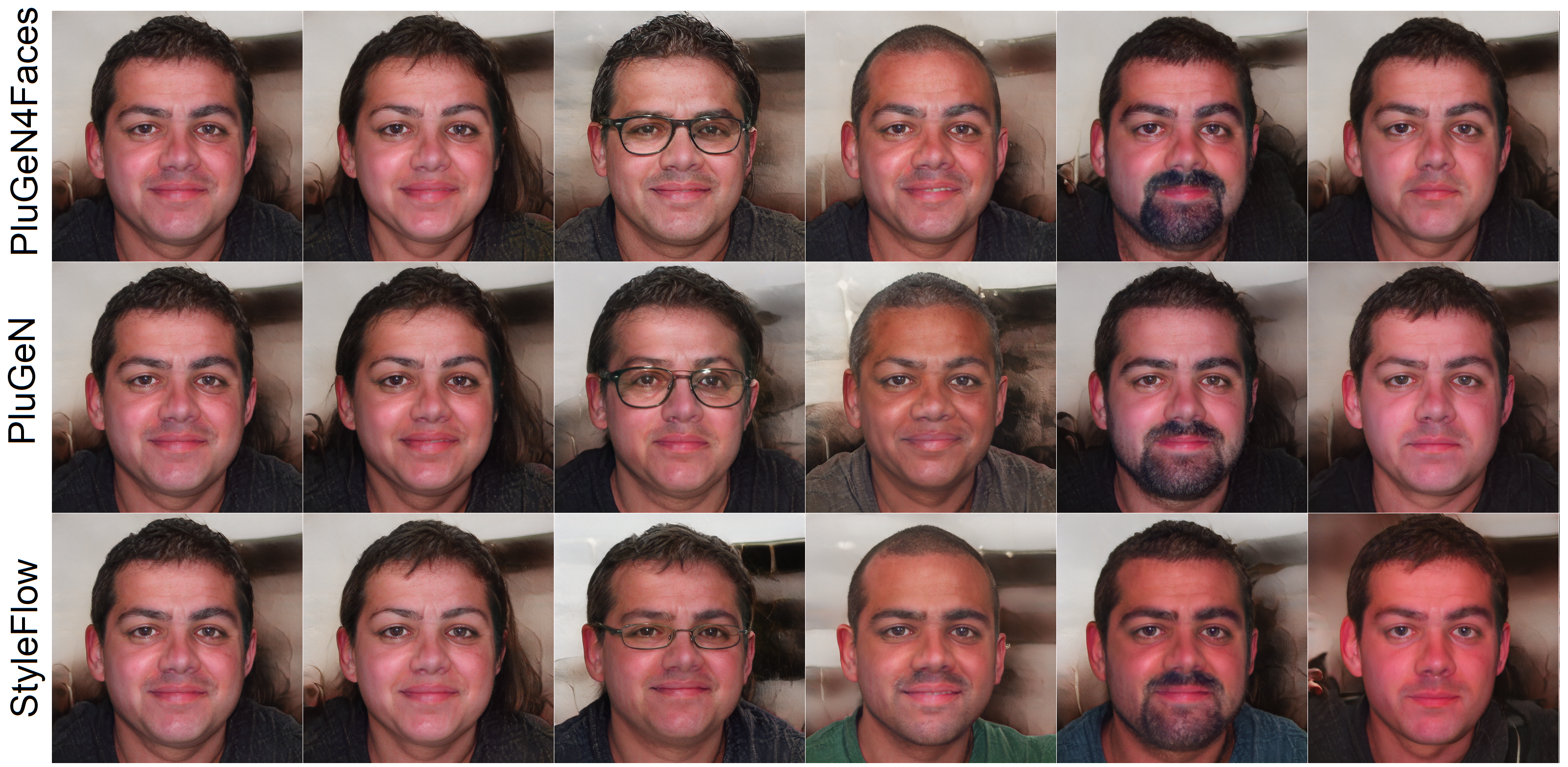} \quad \includegraphics[width=0.48\linewidth]{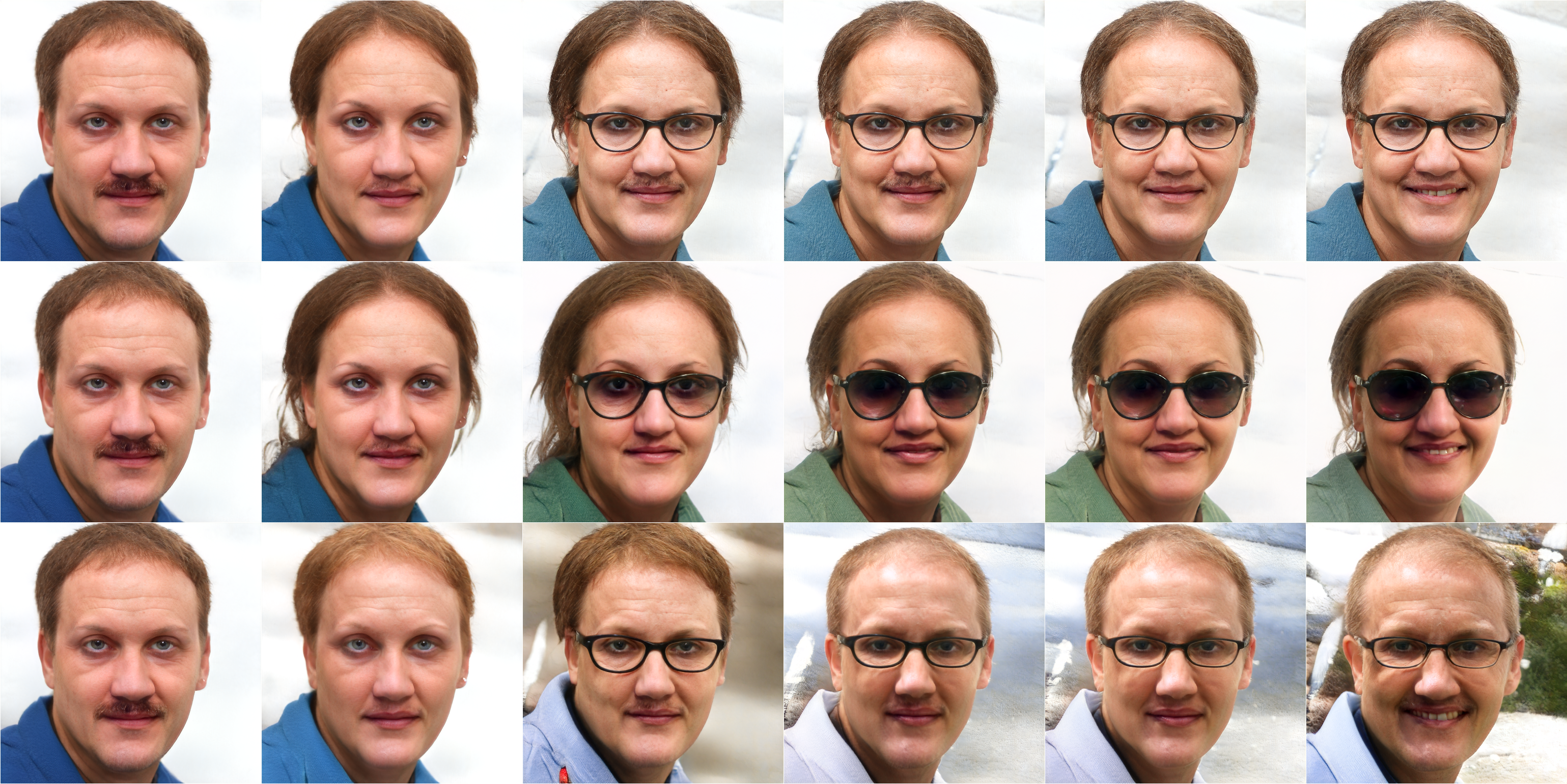}\\ 
    \includegraphics[width=0.48\linewidth]{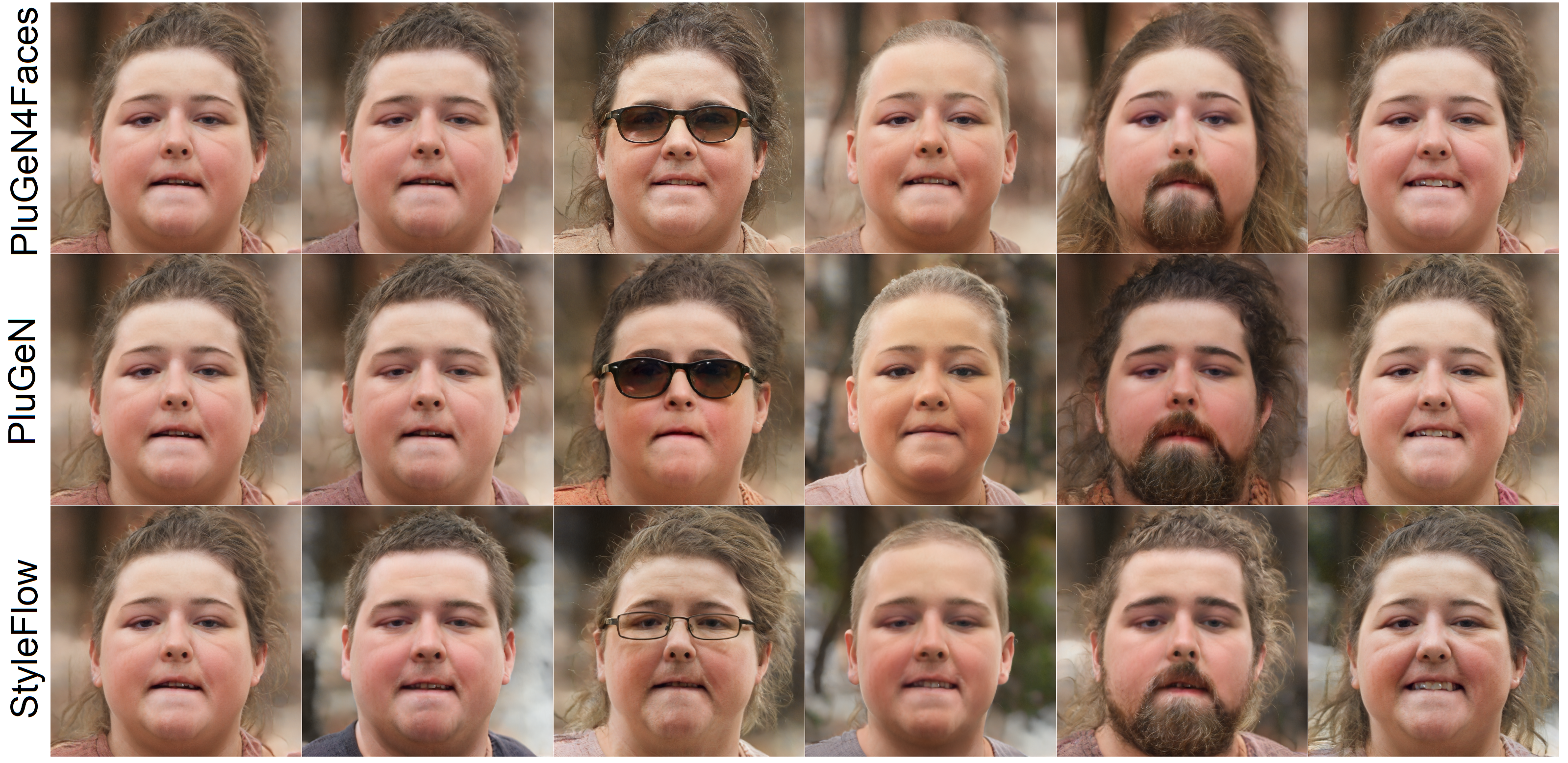} \quad  \includegraphics[width=0.48\linewidth]{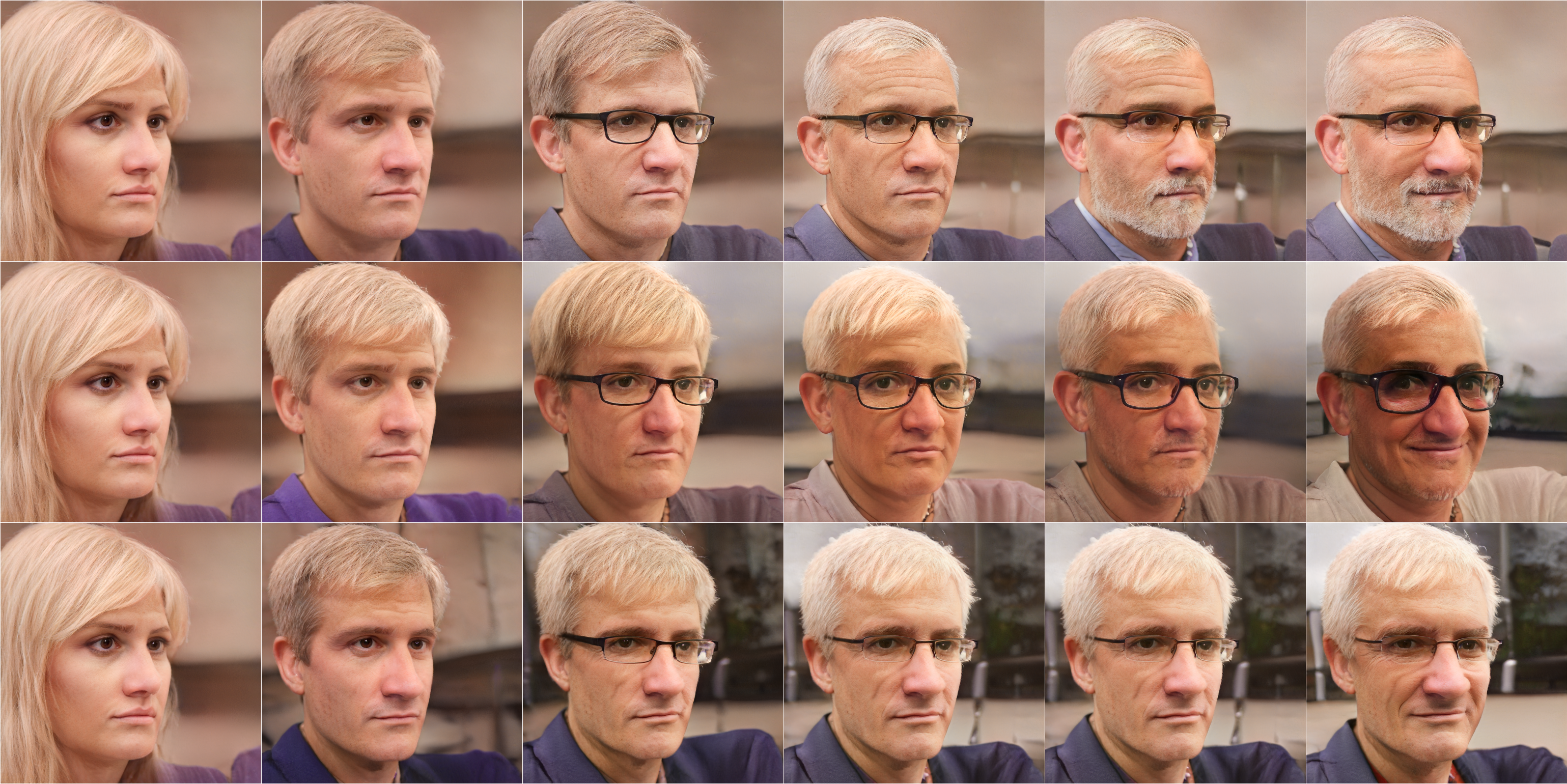}\\
    
    \caption{Single attribute manipulations (left) and sequential edits of multiple attributes (right).}
        \label{fig:modifications}
 \end{figure*}

\paragraph{Inference}

To edit attributes of a real image $\mathbf{x}$, we find its style codes $\mathbf{w}_1,\ldots,\mathbf{w}_k$ using the encoder network. Next, we  map each style code $\mathbf{w}_i$ using the inverse of cINF to obtain the attribute and non-attribute vectors $(\mathbf{c}_i, \mathbf{s}_i)$, for $i=1,\ldots,k$. The requested attributes are modified in each attribute vector $\mathbf{c}_i$ and next they are mapped back by the cINF to the style codes. The synthesis network generates the image with edited attributes from the modified style codes.

\section{Experiments}

\newcolumntype{C}{>{\centering\arraybackslash}X}
  \begin{figure*}[t!]
   \centering
   \begin{tabularx}{\linewidth}{C}{\bf \large Age}  \end{tabularx}\\
    \includegraphics[width=0.49\linewidth]{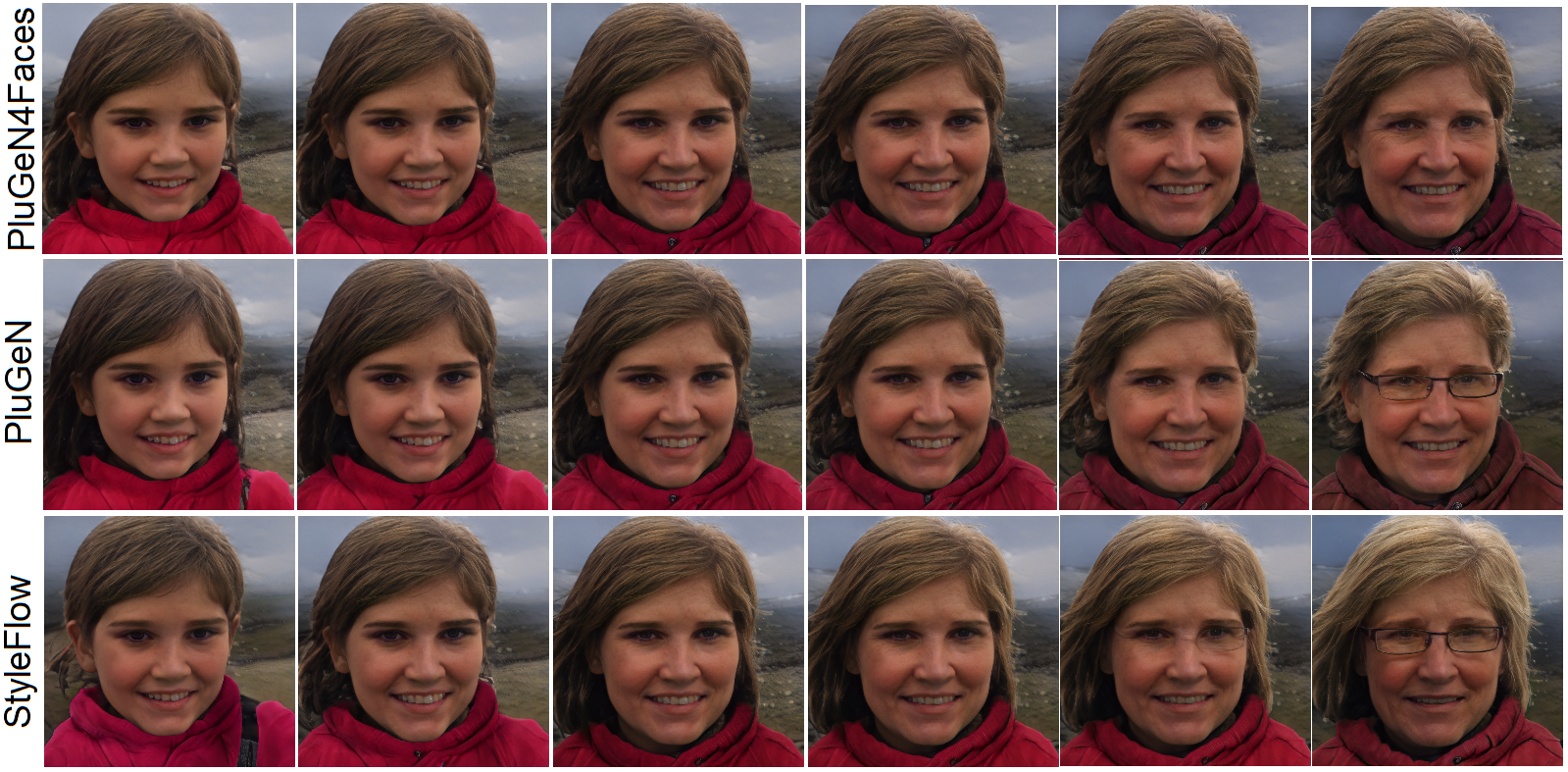} \quad \includegraphics[width=0.48\linewidth]{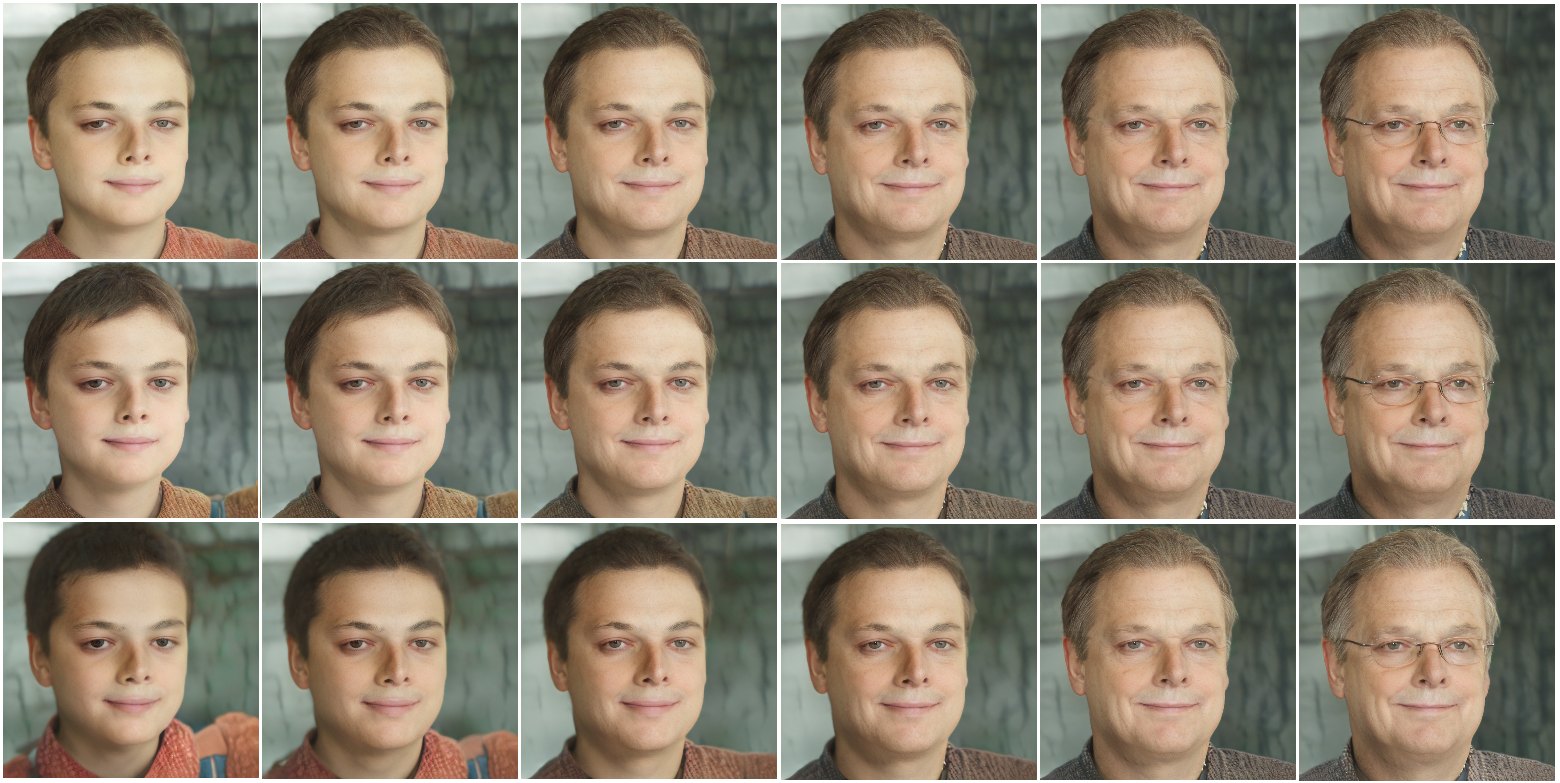}\\
    \begin{tabularx}{\linewidth}{C} {\bf \large Yaw}\end{tabularx} \\
    \includegraphics[width=0.48\linewidth]{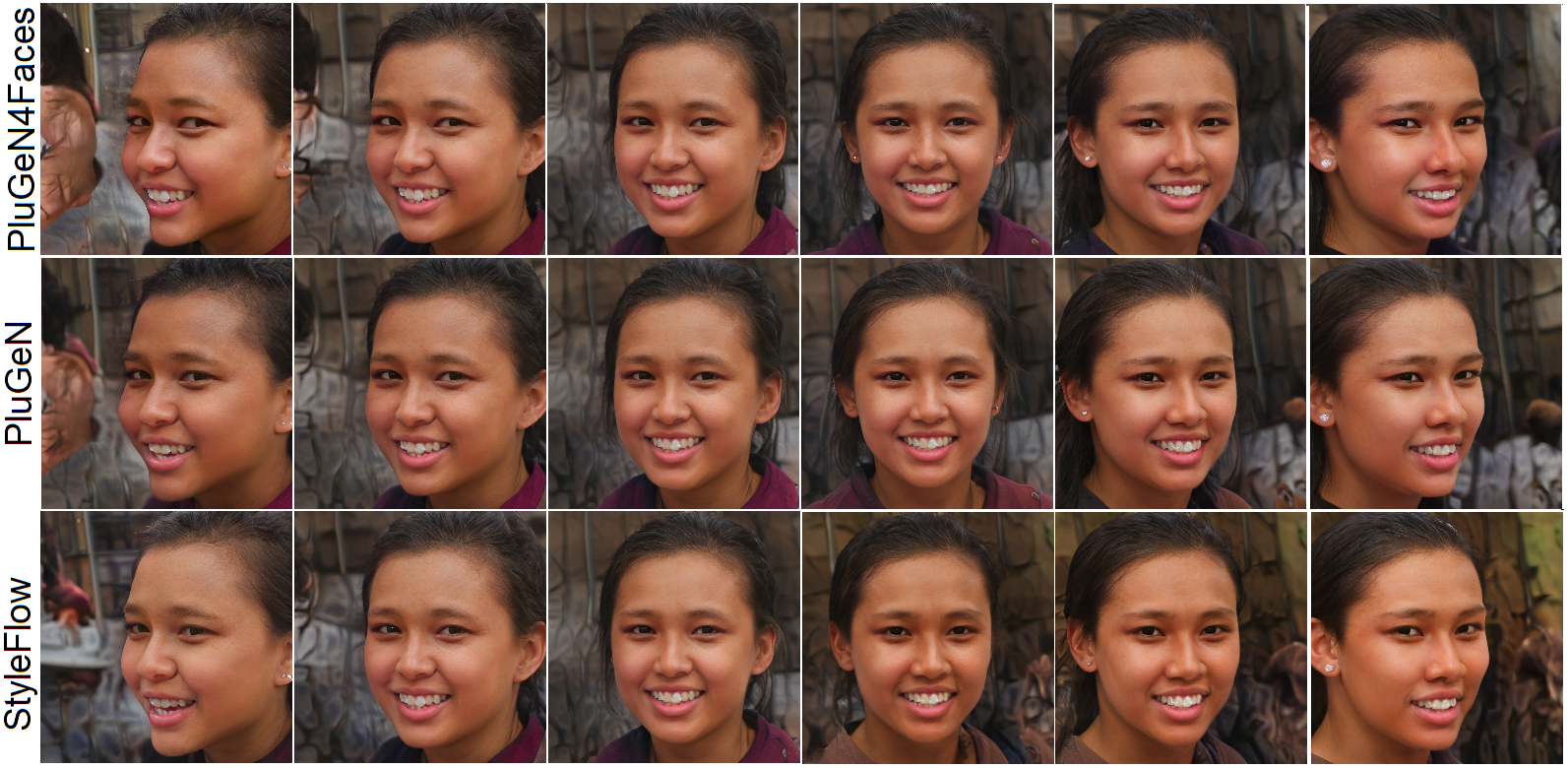} \quad \includegraphics[width=0.48\linewidth]{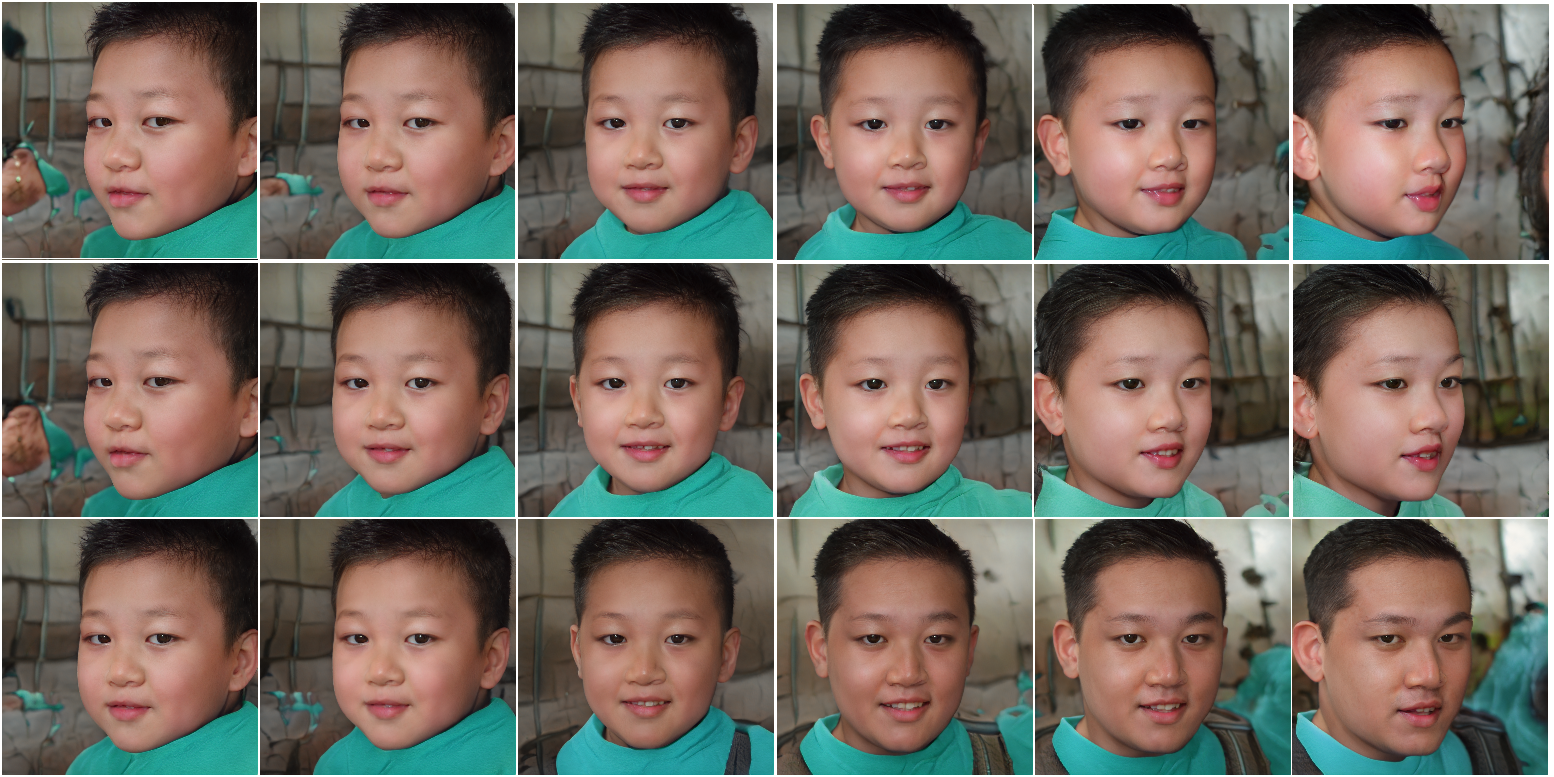}\\
    \begin{tabularx}{\linewidth}{C} {\bf \large Pitch}\end{tabularx} \\
    \includegraphics[width=0.48\linewidth]{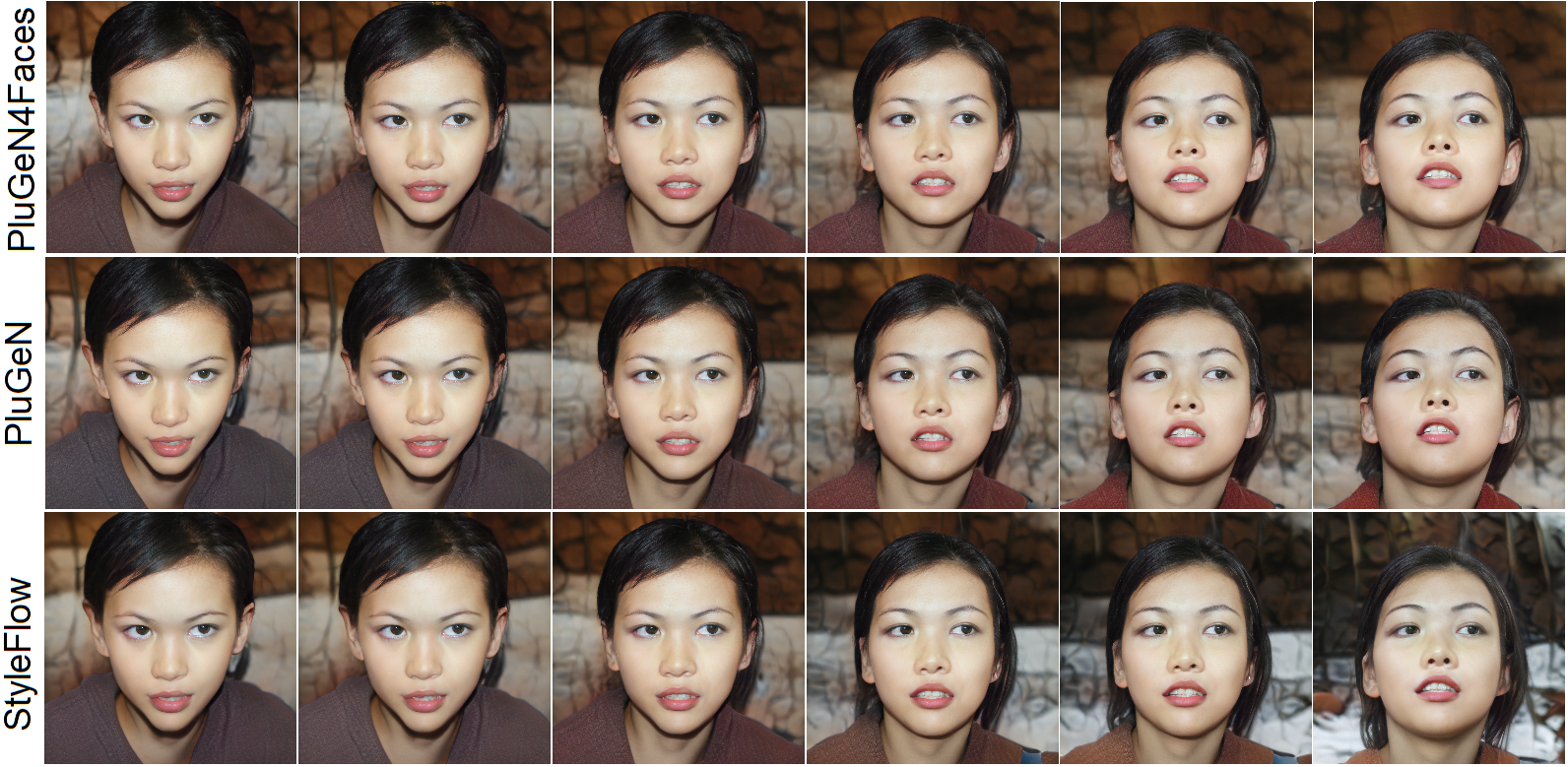} \quad \includegraphics[width=0.48\linewidth]{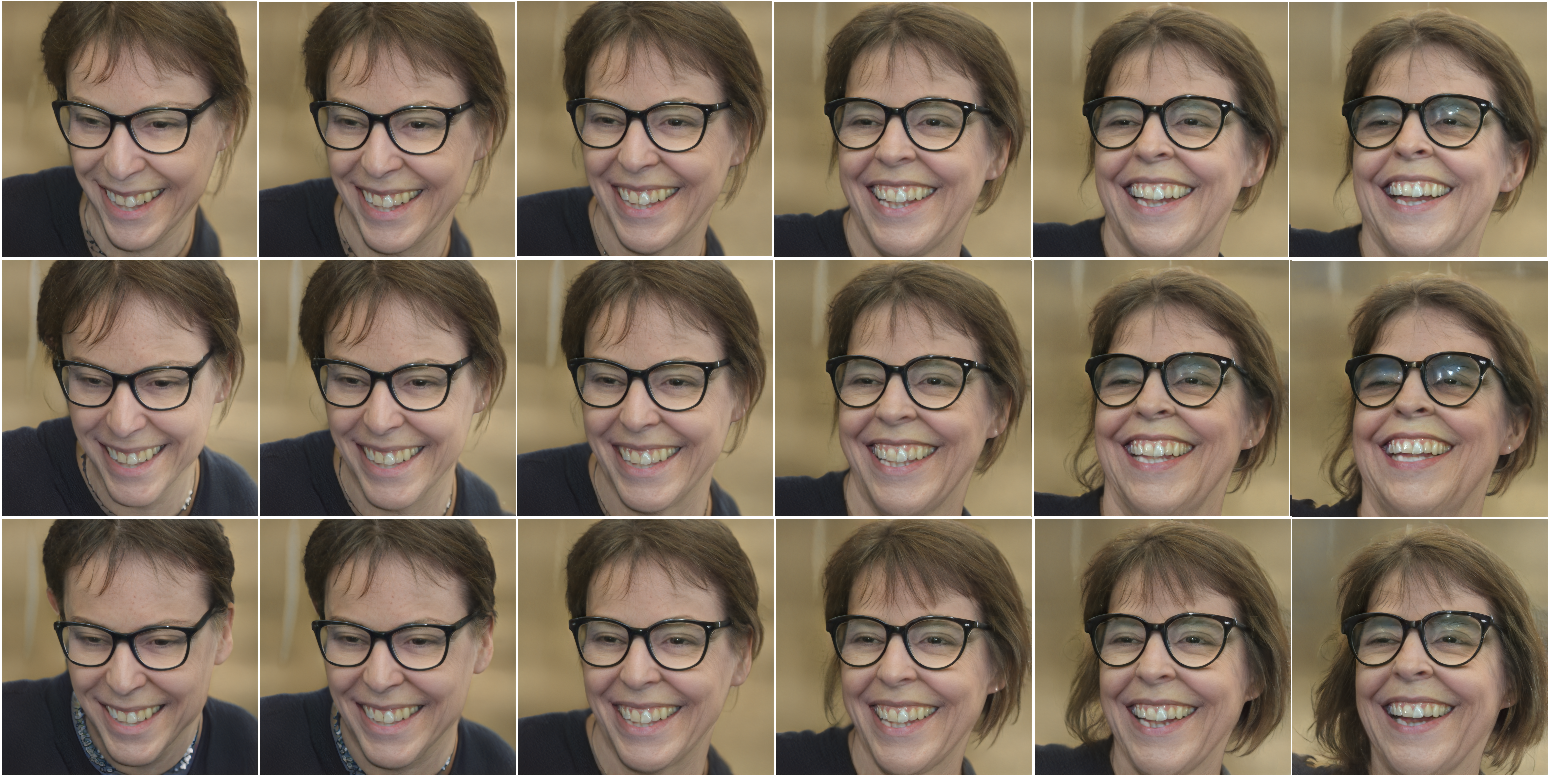}\\
    \caption{Interpolation on the extreme values of continuous attributes.
    \label{fig:continuous}}
 \end{figure*}

\paragraph{Experimental setting}

% \paragraph{Datasets} 
We consider Flickr-Faces-HQ dataset (FFHQ) containing 70 000 high-quality images of resolution $1024 \times 1024$. The Microsoft Face API was used to label 8 attributes in each image (gender, glasses, hair/bald, facial hair/beard, expression/smile, age,  pitch, and yaw). 

Additionally, to explicitly control the person's identity we use images retrieved from video clips. More precisely, we use images from videos and celebrity interviews scraped from YouTube with 573 videos, an average of 19.33 images per video and 12 194 images in total. As in the case of FFHQ dataset, attributes of every image are also labeled using Microsoft Face API. 

% Training is performed using both FFHQ and movie datasets, while evaluation is conducted on images from FFHQ dataset.

% \paragraph{Face Attribute Classifier} 
To evaluate the proposed disentanglement model access to an independent face attribute classifier is needed. For this purpose, we train the ResNet-18 model \citep{he2016deep} on the FFHQ and 10 000 randomly generated StyleGAN face images. The model is trained with 8 outputs in a multi-label manner, treating the Microsoft Face API labels as targets. We standardize the labels as well as apply the shrinkage loss \citep{lu2018deep} as we find that it helps with dataset imbalance. We use the same loss for binary and continuous labels as this works equally well for classification \citep{janocha2016loss}. Although not all of the face attribute labels are binary, we call this model \emph{classifier} in the remainder of this paper to avoid any confusion with the other models used in the experiments.

%\paragraph{Models} 
We use StyleGAN (version 2) as a backbone model, which was trained on FFQH dataset. Real images are encoded to the extended latent space $\W_*^k$ of StyleGAN, where $k=18$, using the encoder network \citep{tov2021designing}. In consequence, every image is represented using a sequence of style codes $\{\mathbf{w}_1,\ldots,\mathbf{w}_k\} \in \W_*^k$, where $\mathbf{w}_i \in \R^{512}$. The encoder is trained on the combination of images from FFHQ and movie datasets.

\our{} is instantiated using conditional RealNVP flow model \citep{dinh2014nice} that operates on the individual latent codes $\mathbf{w}_i \in \R^{512}$ of StyleGAN. The condition is an identifier $i$ of the style code (being the input to the $i$-th StyleGAN layer) represented as a one-hot vector. %As a baseline, we choose \plug{}, a recent state-of-the-art conditional model, which is also used with a pre-trained StyleGAN. \plug{} uses NICE flow model to transform individual style codes $\mathbf{w}_i$ to disentangled space. In contrast to \our{}, \plug{} does not use conditional flow but a single shared flow model (with the same parameters) as a mapping between each style code and the target disentangled space. 

As a baseline, we choose two state-of-the-art conditional models, \plug{} and StyleFlow, which can be used with a pre-trained StyleGAN. \plug{} uses NICE flow model to transform individual style codes $\mathbf{w}_i$ to disentangled space. In other words, \plug{} uses a single shared NICE model (with the same parameters) as a mapping between each style code and the target disentangled space. StyleFlow is parameterized by the conditional continuous flow, where the conditioning factor corresponds to the labeled attributes. Similarly to \plug{}, StyleFlow uses a single flow, which is applied to various style codes. %However, the authors of StyleFlow suggest connecting attributes with a specific number of style codes. More precisely, they empirically found that the attributes are modified using the following indices (layers of the generator): Expression (4-5), Yaw (0-3), Pitch (0-3), Age (4-7), Gender (0-7), Remove Glasses (0-2), Add Glasses (0-5), Baldness (0-5) and Facial hair (5-7 and 10).

%%%%%%%%%%%%%%%%%%%%%%%%%%%%%%%%%%%%

\begin{table*}[!htb]
\centering
\footnotesize
\caption{Identity disentanglement. For each image, we change of the values of attributes listed in rows and compare the relation between original input image and the modified one in terms of 4 measures: (i--ii) MSE between image embedings taken from Face Recogiontion and ArcFace models, (iii) PSNR and (iv) SSIM applied on raw images. \label{fig:ex1} }
\setlength{\tabcolsep}{3pt}
\begin{tabular}{l|cccc|cccc|cccc}
%{p{2cm}|m{1cm}m{1cm}m{1cm}m{1cm}|m{1cm}m{1cm}m{1cm}m{1cm}}
\toprule
 & \multicolumn{4}{c}{{\bf \our{} (ours)}} & \multicolumn{4}{c}{{\bf \plug{}}} & \multicolumn{4}{c}{{\bf StyleFlow}} \\
 \midrule
 & FR & ArcFace & Raw & Raw & FR & ArcFace & Raw & Raw & FR & ArcFace & Raw & Raw\\
 & MSE $\downarrow$ & MSE $\downarrow$ & PSNR $\uparrow$ & SSIM $\uparrow$ & MSE $\downarrow$ & MSE $\downarrow$ & PSNR $\uparrow$ & SSIM $\uparrow$ & MSE $\downarrow$ & MSE $\downarrow$ & PSNR $\uparrow$ & SSIM $\uparrow$\\
 \midrule
male       & 0.20          & \textbf{0.25} & \textbf{30.34} & \textbf{0.84} & \textbf{0.20} & 0.26        & 29.96    & 0.83     & 0.25          & 0.35          & 26.97          & 0.75          \\
female     & \textbf{0.22} & \textbf{0.28} & \textbf{29.58} & \textbf{0.84} & 0.24          & 0.31        & 26.49    & 0.80     & 0.27          & 0.38          & 26.91          & 0.74          \\
glasses    & 0.40          & 0.65          & 20.97          & 0.65          & 0.42          & 0.64        & 19.62    & 0.64     & \textbf{0.36} & \textbf{0.53} & \textbf{22.20} & \textbf{0.66} \\
no glasses & \textbf{0.12} & \textbf{0.11} & \textbf{39.05} & \textbf{0.95} & 0.12          & 0.11        & 37.37    & 0.93     & 0.20          & 0.21          & 27.93          & 0.77          \\
bald       & \textbf{0.14} & \textbf{0.18} & \textbf{29.50} & \textbf{0.82} & 0.22          & 0.27        & 24.32    & 0.74     & 0.21          & 0.28          & 27.45          & 0.72          \\
hair       & \textbf{0.07} & \textbf{0.04} & 38.67          & \textbf{0.95} & 0.10          & 0.07        & 33.19    & 0.90     & 0.10          & 0.09          & \textbf{38.77} & 0.88          \\
old        & 0.45          & \textbf{0.67} & \textbf{22.75} & \textbf{0.66} & 0.45          & 0.72        & 20.65    & 0.62     & \textbf{0.45} & 0.70          & 20.63          & 0.57          \\
young      & \textbf{0.43} & \textbf{0.63} & \textbf{22.71} & \textbf{0.69} & 0.46          & 0.75        & 20.61    & 0.63     & 0.43          & 0.73          & 21.40          & 0.60          \\
beard      & 0.29          & 0.35          & 23.54          & 0.75          & 0.33          & 0.47        & 21.25    & 0.67     & \textbf{0.21} & \textbf{0.23} & \textbf{31.18} & \textbf{0.80} \\
no beard   & \textbf{0.10} & \textbf{0.07} & \textbf{39.58} & \textbf{0.94} & 0.11          & 0.09        & 35.11    & 0.91     & 0.15          & 0.15          & 32.10          & 0.83          \\
smile      & \textbf{0.11} & \textbf{0.07} & \textbf{35.75} & \textbf{0.93} & 0.14          & 0.10        & 29.87    & 0.86     & 0.17          & 0.16          & 29.83          & 0.79          \\
no smile   & 0.19          & 0.16          & 29.63          & \textbf{0.86} & 0.22          & 0.21        & 24.31    & 0.74     & \textbf{0.17} & \textbf{0.15} & \textbf{30.96} & 0.81          \\
up         & \textbf{0.22} & \textbf{0.23} & 24.58          & \textbf{0.76} & 0.24          & 0.27        & 22.30    & 0.71     & 0.26          & 0.35          & \textbf{25.32} & 0.67          \\
down       & \textbf{0.16} & \textbf{0.14} & 28.63          & \textbf{0.84} & 0.18          & 0.17        & 26.13    & 0.80     & 0.18          & 0.22          & \textbf{33.44} & 0.78          \\
right      & \textbf{0.25} & \textbf{0.32} & 19.88          & \textbf{0.60} & 0.25          & 0.32        & 19.30    & 0.59     & 0.29          & 0.41          & \textbf{23.72} & 0.55          \\
left       & \textbf{0.22} & \textbf{0.27} & 21.58          & \textbf{0.65} & 0.22          & 0.27        & 20.88    & 0.64     & 0.26          & 0.36          & \textbf{26.64} & 0.60          \\
\midrule
avg        & \textbf{0.22}          & \textbf{0.28}          & \textbf{28.54}          & \textbf{0.79}          & 0.24          & 0.31        & 25.71    & 0.75     & 0.25          & 0.33          & 27.84          & 0.72        \\
\bottomrule
\end{tabular}
\end{table*}

\paragraph{Qualitative results}

In this section, we illustrate the sample results produced by the proposed model. First, we perform a single edit of binary attributes. Next, we consider sequential edits, where subsequent modifications on binary attributes are added one by one. In both cases, we perform a minimal modification needed to change a decision of the attribute classifier. More precisely, we perform a gradual change of the attribute and inspect the reaction of the attribute classifier on the modified attribute of the generated image. If the classifier recognizes the attribute of the generated image with sufficient confidence, then we stop modification and return the generated image. By making use of an independent classifier, we are guaranteed to obtain a fair comparison regardless of the scale used by the models. %To realize this procedure effectively, we perform a binary search on the latent representation of the attribute. The search range and the classifier confidence threshold are individually selected for a given attribute.

Figure \ref{fig:modifications} presents the results of single (left) and sequential edits (right). At first glance, all considered models give visually appealing effects and perform successfully the requested modifications. Observe however that \plug{} and StyleFlow changed the ethnicity and age of the person in the top left example when modifying the attribute "hair". Such a behavior is not accepted and does not hold in the case of \our{} (see the 1st row of Figure \ref{fig:intro}, where \our{} added hair without changing the ethnicity). It is impressive that all models were able to combine the attribute "beard" with a woman's face in the bottom left example. Nevertheless, the face produced by \our{} has more female features than the ones generated by \plug{} and StyleFlow. Looking at sequential edits (right), it is evident that \our{} kept the color of clothes and background unchanged, which is not the case of \plug{} and StyleFlow. Moreover, the type of glasses is also unaffected by attribute manipulations performed by \our{}. On the downside, it should be noted that all models make the face slightly older when the attributes "bald" or "beard" are used.

We also illustrate the manipulations of continuous attributes by showing the path between two extreme values of a given attribute, see Figure \ref{fig:continuous}. Although the requested modifications have been successfully realized by the models, \our{} was less invasive to the images. \plug{} could not avoid adding glasses when changing the age (left); it modified the gender of the child's face when turning the head left (middle right); it changed the color of clothes in the bottom left example. StyleFlow modified the age of a child when turning his head right (middle right) as well as added male features to the face presented in the bottom right example when the head was turned down. \our{} was free of the aforementioned drawbacks, which demonstrates that it better disentangles the image space and is able to preserve more of the original features during edits.

%%%%%%%%%%%%%%%%%%%%%%%%%%%%%%%%%%%%%%

\begin{table}[!htb]
\centering
\caption{Attributes disentanglement measured by the accuracy (higher is better). For each image, we change of the values of attributes listed in rows and verify whether the remaining attributes (listed in columns) stay unchanged. We report the percentage of successes (accuracy). In the last column, we also report the accuracy of modifying the requested attribute (listed in rows).    \label{tab:ex2} }
\footnotesize
\setlength{\tabcolsep}{3pt} % Default value: 6pt
\begin{tabular}{l|ccccc|c|c}
\toprule
  % & smile          & gender         & glasses        & bald           & beard          & avg.            & acc. of          \\
  & gender          & glasses         & bald        & beard           & smile          & avg.            & acc. of          \\
       &           &          &         &            &           &             & modif.            \\
                              \midrule
    & \multicolumn{7}{c}{{\bf \our{} (ours)}}\\
\midrule
 gender  & -      & 96.99   & \textbf{90.90} & 85.75 & 89.27 & 90.72 & \textbf{91.94}         \\
                               glasses & \textbf{95.25}  & -       & 92.01 & 86.69 & 89.48 & 90.86 & 99.10         \\
                               bald    & \textbf{94.79}  & 97.17   & -     & \textbf{86.98} & \textbf{90.23} & \textbf{92.29} & \textbf{96.19}         \\
                               beard   & \textbf{94.92}  & 96.46   & \textbf{93.41} & -     & \textbf{90.75} & \textbf{93.88} & 66.91         \\
                               smile   & \textbf{95.84}  & \textbf{96.13}   & 93.41 & 86.86 & -     & \textbf{93.06} & \textbf{98.14}         \\
                                avg.       &        &         &       &       &       & \textbf{92.16} & \textbf{90.46}         \\
                                \midrule
    & \multicolumn{7}{c}{{\bf \plug{}}}\\
                              \midrule
gender  & -      & \textbf{97.70}   & 90.69 & \textbf{85.81} & 89.87 & \textbf{91.02} & 84.28         \\
                               glasses & 93.28  & -       & 92.57 & 86.77 & 89.68 & 90.58 & \textbf{99.41}         \\
                               bald    & 93.74  & \textbf{97.20}   & -     & 86.48 & 89.87 & 91.82 & 72.37         \\
                               beard   & 86.82  & \textbf{97.14}   & 93.03 & -     & 90.34 & 91.83 & 75.93         \\
                               smile   & 92.17  & 96.05   & \textbf{93.45} & 86.75 & -     & 92.10 & 97.28         \\
                                avg.       &        &         &       &       &       & 91.47 & 85.86         \\
                                \midrule
    & \multicolumn{7}{c}{{\bf StyleFlow}}\\
                              \midrule
 gender  & -      & 95.38   & 90.46 & 85.65 & \textbf{90.23} & 90.43 & 90.52         \\
                               glasses & 94.48  & -       & \textbf{92.82} & \textbf{87.09} & \textbf{90.42} & \textbf{91.20} & 98.70         \\
                               bald    & 91.86  & 95.46   & -     & 86.77 & 87.32 & 90.35 & 73.80         \\
                               beard   & 83.47  & 95.80   & 92.59 & -     & 89.70 & 90.39 & \textbf{77.65}         \\
                               smile   & 94.92  & 96.11   & 93.39 & \textbf{87.34} & -     & 92.94 & 76.04         \\
                               avg.        &        &         &       &       &       & 91.06 & 83.34        \\
                              \bottomrule
\end{tabular}
\end{table}

\begin{table}[!htb]
\caption{Attributes disentanglement measured by the ranking correlation (higher is better). For each image, we change the values of attributes listed in rows and verify whether the ranking of the remaining attributes (listed in columns) given by the classifier outputs stay unchanged. We report the correlation between rankings before and after the change.  \label{tab:ex3}}
\footnotesize
\setlength{\tabcolsep}{3pt}
\begin{tabular}{l|cccccccc}
% \begin{tabularx}{\textwidth} { 
%   p{0.8cm} | p{1.8cm} | >{\centering\arraybackslash}X >{\centering\arraybackslash}X >{\centering\arraybackslash}X >{\centering\arraybackslash}X
%  >{\centering\arraybackslash}X >{\centering\arraybackslash}X >{\centering\arraybackslash}X >{\centering\arraybackslash}X }
  \toprule
& gender         & glasses        & bald           & beard          & smile          & age            & pitch          & yaw            \\
\midrule
    & \multicolumn{8}{c}{{\bf \our{} (ours)}}\\
\midrule
 gender  &         -       & \textbf{87.01} & \textbf{90.91} & 78.83          & 95.17          & 96.53          & \textbf{98.92} & \textbf{99.79} \\
                              glasses & \textbf{93.65} &        -        & \textbf{91.51} & \textbf{96.15} & \textbf{95.20} & 95.83          & \textbf{98.31} & \textbf{99.79} \\
                              bald    & \textbf{93.98} & \textbf{89.18} &        -        & \textbf{96.17} & \textbf{96.87} & \textbf{98.68} & \textbf{99.05} & \textbf{99.75} \\
                             beard   & \textbf{90.81} & \textbf{86.97} & \textbf{91.49} &         -       & \textbf{94.48} & 96.58          & 98.54          & \textbf{99.71} \\
                              smile   & \textbf{95.50} & \textbf{88.61} & \textbf{95.53} & \textbf{96.86} &       -         & \textbf{98.38} & \textbf{98.96} & {\bf 99.74}          \\
                             age     & \textbf{86.02} & \textbf{39.34} & \textbf{83.36} & \textbf{87.94} & \textbf{89.28} &    -            & 95.50          & \textbf{99.58} \\
                             pitch   & \textbf{95.57} & \textbf{89.25} & \textbf{94.45} & \textbf{96.94} & \textbf{96.25} & \textbf{98.93} &      -          & \textbf{99.82} \\
                              yaw     & 91.41          & 83.49          & 90.59          & 93.66          & 93.35          & 96.96          & 97.60          &        -        \\
                              \midrule
avg     & \textbf{92.42} & \textbf{80.55} & \textbf{91.12} & \textbf{92.36} & \textbf{94.37} & 97.41          & \textbf{98.12} & \textbf{99.74} \\
\midrule
    & \multicolumn{8}{c}{{\bf \plug{}}}\\
\midrule
 gender  &       -         & 86.86          & 89.73          & \textbf{79.93} & \textbf{95.56} & \textbf{96.84} & 98.53          & 99.65          \\
                              glasses & 92.52          &         -       & 91.14          & 95.44          & 94.09          & \textbf{96.00} & 97.74          & 99.64          \\
                             bald    & 92.95          & 87.07          &      -          & 95.14          & 95.33          & 98.11          & 98.73          & 99.60          \\
                              beard   & 85.43          & 85.65          & 88.66          &     -           & 93.41          & \textbf{97.21} & \textbf{98.57} & 99.47          \\
                              smile   & 90.66          & 85.87          & 94.02          & 93.73          &      -          & 98.19          & 98.51          & 99.59          \\
                             age     & 80.06          & 38.11          & 76.66          & 79.38          & 89.00          &      -          & \textbf{96.13} & 99.44          \\
                              pitch   & 94.47          & 85.30          & 94.02          & 96.41          & 95.84          & 98.62          &       -         & 99.74          \\
                              yaw     & \textbf{92.42} & \textbf{84.31} & \textbf{92.48} & \textbf{95.18} & \textbf{94.62} & \textbf{98.32} & \textbf{98.01} &       -         \\
                              \midrule
 avg     & 89.78          & 79.02          & 89.53          & 90.74          & 93.97          & \textbf{97.61} & 98.03          & 99.59          \\
\midrule
    & \multicolumn{8}{c}{{\bf StyleFlow}}\\
\midrule
 gender  &        -        & 80.42          & 87.13          & 65.90          & 94.37          & 95.64          & 97.76          & 99.42          \\
                              glasses & 91.11          &           -     & 90.69          & 93.99          & 93.48          & 95.03          & 97.65          & 99.46          \\
                              bald    & 89.72          & 83.20          &        -        & 93.86          & 92.88          & 97.28          & 97.97          & 99.03          \\
                              beard   & 80.51          & 84.55          & 89.33          &      -          & 93.80          & 95.89          & 97.84          & 99.08          \\
                              smile   & 92.84          & 86.65          & 92.73          & 95.57          &          -      & 97.81          & 98.30          & 99.62 \\
                              age     & 82.13          & 34.16          & 75.60          & 80.92          & 88.34          &    -            & 93.24          & 98.74          \\
                              pitch   & 90.44          & 82.07          & 91.69          & 94.46          & 94.15          & 97.76          &       -         & 99.51          \\
                               yaw     & 87.72          & 78.93          & 86.40          & 92.49          & 91.91          & 95.45          & 95.53          &      -          \\
                              \midrule
avg     & 87.78          & 75.71          & 87.65          & 88.17          & 92.70          & 96.41          & 96.90          & 99.26         \\
\bottomrule
% \end{tabularx}
\end{tabular}
\end{table}

\paragraph{Identity preservation}

In this part, we support our sample results with quantitative evaluation, which aims at verifying how well \our{} disentangles the image representation. To this end, we change a single attribute of a given image and compare the resulting picture with the original image (before modification). Again, for a fair comparison, we employ a classifier and apply a minimal modification which is accepted by the attribute classifier.

To compare the difference between images, we apply two approaches. In the first one, we calculate the mean square error (MSE) between embeddings of the original and modified images taken from a pre-trained network. To this end, we employ two networks applicable to processing face images: ArcFace\footnote{\url{https://github.com/deepinsight/insightface}} \citep{deng2019arcface} and FR\footnote{\url{https://github.com/ageitgey/face_recognition}}. A model with a lower MSE preserves more features (including identity) from the original image. Second, to explicitly compare the difference between images we also use the PSNR and SSIM measures applied to raw images. Such measures suit perfectly to compare the modification of low-level features such as the background. 

Table \ref{fig:ex1} shows how the proposed measures react to changing subsequent face attributes. Each row corresponds to the requested value of the modified attribute. The results consistently confirm that \our{} obtains significantly better scores than \plug{} and StyleFlow in most cases. One can observe that modifying the "age" attribute has a significant effect on the disentanglement measures, which suggests that changing the age leads to changes in a person's identity. It is interesting that modifying gender in face images has a moderate influence on face identification. This could mean that both models successfully disentangled this attribute from the remaining image information. The smallest changes are observed for manipulating "smile" and "hair" attributes.

\paragraph{Attributes disentanglement}

We also verify the disentanglement between labeled attributes in a strict quantitative way. Namely, we force the change of a single attribute and verify whether the values of other labeled attributes changed as well. Ideally, the values of the remaining attributes should stay intact.

For binary attributes (smile gender, glass, hair, and beard), we apply a standard accuracy measure, which shows whether the classifier keeps its original prediction on non-modified attributes. Additionally, we employ a ranking measure, which can be used for discrete as well as continuous attributes because classifier scores do not have to be discretized in this case. %The ranking measure compares if the ranking produced by the classifier on any attribute remains the same after modifying a given attribute. More precisely, w
In this approach, we rank input (non-modified) images based on the scores returned by the classifier on the attribute $A_i$. Next, we change the value of the attribute $B$ and again calculate the ranking using the classifier scores based on the attribute $A_i$. We compare the rankings before and after the change using the Rank Correlation Coefficient (Spearman's $\rho$), which gives a maximal value of 1, for two identical rankings. Higher values indicate better disentanglement. We repeat this experiment for all attributes $A_1,\ldots, A_k$. 

Table \ref{tab:ex2} shows that all models obtain the average accuracy on non-target attributes above 90\% and around 80\% on the attributes being modified, which means that it is still more difficult to perform the modification than to keep the values of other features. Taking the average of accuracy scores reveals that \our{} outperforms \plug{} and StyleFlow in both metrics. Looking at the ranking correlation presented in Table \ref{tab:ex3}, we observe that the advantage of \our{} over \plug{} and StyleFlow is even higher. It gives higher scores in 41 out of 56 cases. 

The lowest correlation scores were obtained when we modified the age attribute (which aligns with the conclusion of the previous experiment). It was almost impossible to keep the ranking on the glasses attribute, which might be explained by the fact that the training does not contain young people wearing glasses. Previous sample results presented in Figure \ref{fig:continuous} also showed that increasing the age attribute accidentally leads to adding glasses. Analogical negative behavior occurs in the case of beard and hair attributes, which are highly correlated with age. This analysis shows that it is very difficult to overcome the bias introduced in a training set and provide high-quality disentanglement between some face attributes.

\section{Conclusion}

We introduced \our{} for disentangling face attributes from the person's identity. The proposed model works as a plugin to the pre-trained StyleGAN model, which makes it extremely easy to use in practice. Our key idea relies on applying contrastive learning on images retrieved from movie frames that contain information about a person's identity. Our experiments supported by the rigorous quantitative analysis demonstrate that \our{} is focused on manipulating the requested attributes and is less invasive to the remaining image attributes than the existing methods.

%%%%%%%%% REFERENCES
{\small
\bibliographystyle{plain}
\bibliography{ref.bib}
}

\end{document}